\documentclass{article}

\usepackage{booktabs} 
\usepackage{amsmath}
\usepackage{amsfonts}
\usepackage{graphicx}
\usepackage[utf8]{inputenc}
\usepackage{natbib}

\begin{document}
\title{A Reinforcement Learning Perspective on the Optimal Control of Mutation Probabilities for the (1+1) Evolutionary Algorithm: First Results on the OneMax Problem}

\author{Luca Mossina$^1$ \and Emmanuel Rachelson$^1$ \and Daniel Delahaye$^2$ }
\date{%
    $^1$ISAE-SUPAERO, Universit\'{e} de Toulouse\\%
    \texttt{name.surname@isae-supaero.fr}\\%
    $^2$ENAC, Universit\'{e} de Toulouse\\[2ex]%
}

\maketitle

\begin{abstract}
We study how Reinforcement Learning can be employed to optimally control parameters in 
evolutionary algorithms.
We control the mutation probability of a (1+1) evolutionary algorithm on the OneMax function.
This problem is modeled as a Markov Decision Process and solved with Value Iteration via the known transition probabilities.
It is then solved via $Q$-Learning, a Reinforcement Learning algorithm, where the exact transition probabilities are not needed.
This approach also allows previous expert or empirical 
knowledge to be included into learning.
It opens new perspectives, both formally and computationally, for the problem of parameter control in optimization.
\end{abstract}

\section{Problem statement}
We maximize the \textit{OneMax} function: $OM(x) = \sum_{i=1}^{n}x_i, \forall x_i \in \{0,1\}$ via the (1+1) Evolutionary Algorithm (EA) by which, given a random initialization of $x \in \{0,1\}^n$, at every iteration, each of the bits is flipped (\textit{mutated}) with probability $\theta$, yielding a solution candidate $x'$.
If $OM(x') > OM(x)$, $x'$ is kept.
We proceed until the terminal condition $OM(x) = n$ is met.
We regard the evolution of $x$ as a stochastic process, conditioned at each step by $\theta$. This yields a Markov Decision Process (MDP \citep{puterman2014markov}), whose optimal control policy can be found via Dynamic Programming when the transition probabilities are known and Reinforcement Learning (RL) when only experience data is available\footnote{An online compendium with proofs and code to replicate our results is available at https://github.com/**********}.

\section{Related Work}
Recent results \citep{karafotias2015parameter} have proposed new mechanisms to dynamically control parameters in evolutionary algorithms, in opposition to just tuning and fixing them prior to optimization.
Some theoretical results \citep{bottcher2010optimal, doerr2018effectiveness, Doerr_2015_Optimal_Parameter, giessen_2015_population} have demonstrated the intuition (e.g. 1/5th rule) that adaptive parameters can perform substantially better than static tuning, producing also optimal behaviours in some cases.
When exact analyses are not possible, we propose to use RL \citep{sutton1998reinforcement} to estimate such optimal behaviours.
Indeed, promising results \citep{karafotias2014Generic_paco_rl, budzalova2014selecting} have hinted the potential of the generic use of RL in EA.
\section{Markov Decision Process}
During the execution of the EA, we want to sequentially change $\theta$ to minimize the expected termination time.
This problem can be formulated as an MDP, with states $S = \{0, 1, 2, \dots, n\}, s = OM(x)$ and actions $A = \{\theta_1, \theta_2, \dots\} = \{0.01, 0.02, \dots, 0.99, 1\}$ (a discretization of the mutation probability $\theta \in [0,1]$).
At each step, a reward $r(s,\theta)$ is obtained, where $r(s,\theta) = 0$ if the terminal state $s=n$ is reached, and $r(s,\theta)=-1$ otherwise.
An optimal parameter control policy $\pi(s)=\theta$ maximizes\footnote{For readers used to MDP notations: this total reward criterion (no discount factor $\gamma$) is well defined for Stochastic Shortest Path problems such as the one considered here.} $\mathbb{E}\left( \sum_{t=0}^{\infty} r_{t} \right)$ for any initial state $s$.

\subsection{Transition Probabilities}
The transition matrix $P = [\mathbb{P}\left(s'\mid s, \theta \right)]_{\forall (s,\theta) \in (S, A)}$, describes the probability of transitioning to a state $s'=OM(x')$ from any $s=OM(x)$ given any action $\theta \in A$.
At any iteration $t$, $x_t \in \{0,1\}^n$ has $OM(x_t) = n_1$ ones and $n_0 = n - n_1$ zeros.
Let $W \sim Bin(n_0, \theta)$\footnote{binomial distribution of parameters $(n_0,\theta)$.} be the random variable (r.v.) describing the ones gained at the end of an iteration and $L \sim Bin(n_1, \theta)$ be the r.v. for the ones lost.
$Z' = W - L \in [-n_1, -n_1+1, \dots, n_0 - 1, n_0]$ is the r.v. for the net gain after a mutation.
Note that $Z' = W - L$ is the difference of independent binomial distributions.
By convolution, it follows that the probability mass function of $Z'$ is:
\[  p_{Z'}(z)=\left\{
              \begin{array}{ll}
                \sum_{i = 0}^{n_0} p_W(i+z; n_0, \theta) \times p_L(i  ; n_1, \theta): z \geq 0 \\
                \sum_{i = 0}^{n_1} p_W(i  ; n_0, \theta) \times p_L(i+z; n_1, \theta): z < 0.
              \end{array}
            \right.
\]
Under the (1+1)EA, if $Z' \leq 0$, the solution candidate $x'$ is rejected as no negative values are admissible.
The r.v. $Z'_s$ for the state $s'$ of our EA process has thus values $Z'_s \in [0, 1, 2, \dots, n_0 - 1, n_0]$, where $\mathbb{P}(Z'_s = 0) = \mathbb{P}(Z' <0) + \mathbb{P}(Z'=0)$ and $\mathbb{P}(Z'_s = k) = \mathbb{P}(Z'=k) \ \forall k > 0$.

\section{Optimal Parameter Control}
We briefly introduce the two main methods used to compute the optimal policy: one based on Dynamic Programming \citep{bellman1957dynamic}, the other relying on $Q$-Learning \citep{watkins1989learning}.

\subsection{Dynamic Programming}
The function $V^\pi:s\mapsto \mathbb{E}\left( \sum_{t=0}^{\infty} r_{t} | s_0=s\right)$ (called $\pi$'s value function) maps state $s$ to (minus) their expected time-to-termination. The optimal policy's value function $V^*$ is defined recursively by Equation \ref{eq:VI}. 
Value Iteration is the Dynamic Programming algorithm that repeatedly applies Equation \ref{eq:VI} until convergence to $V^*$.
\begin{equation}
V_{n+1}(s) = \max_\theta \left[r(s,\theta) + \sum_{s'} \mathbb{P}(s'|s,\theta) V_n(s') \right]
\label{eq:VI}
\end{equation}

\begin{figure}
	\centering
	\includegraphics[width=0.7\linewidth]{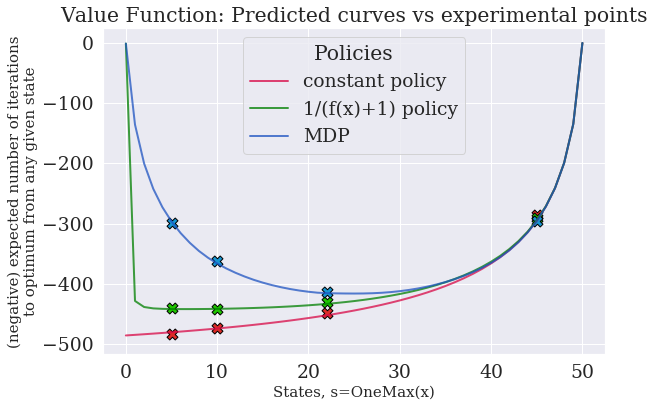}
	\caption{Value function}
	\label{fig:value_f}
\end{figure}

\begin{figure}
	\centering
	\includegraphics[width=0.7\linewidth]{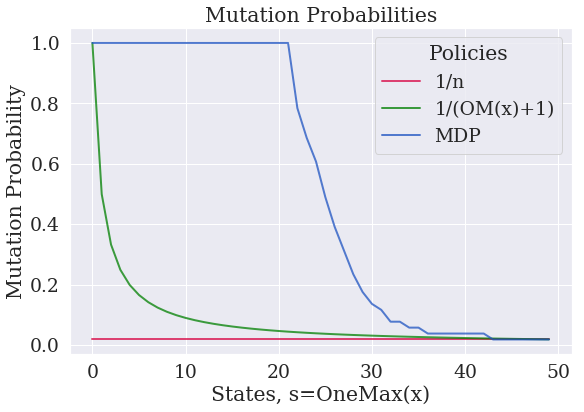}
	\caption{Mutation policies}
	\label{fig:muts}
\end{figure}

Figure \ref{fig:value_f} reports the computed value functions for the following three policies (plotted in Figure \ref{fig:muts}):
\begin{itemize}
	\item the constant $\theta=\frac{1}{n}$ commonly used in (1+1)EA,
	\item the $\pi(s) = \frac{1}{1+s}$ policy from \citep{bottcher2010optimal} (originally designed for the LeadingOnes function),
	\item the optimal policy found via Value Iteration.
\end{itemize}

In Figure \ref{fig:value_f} the marks corresponds to the empirical average $T$, for 2000 runs initialized respectively at $S_{init}= \{5, 10, 22, 45\}$.
In Table \ref{tab:perfo} one can find the average $T$ for a random starting state.

\begin{table}
	\centering
	\small
\begin{tabular}{r|ccc}
Policy                & Constant             & 1/(s+1)              & MDP                  \\ \hline
Average               & 442                  & 430                  & 412                  \\
Standard Deviation    & 163                  & 165                  & 164                  
\end{tabular}
\caption{Empirical time-to-termination, 2000 runs}
\label{tab:perfo}
\end{table}

\subsection{$Q$-Learning}
Although one can explicitly compute the transition probabilities for the parameter control problem based on the OneMax function, such probabilities are generally not available. 
Learning mechanisms such as $Q$-Learning, allow to obtain $\pi^*$, using sampled transitions, without explicitly requiring $P$. 

\begin{equation}
Q(s,\theta) \leftarrow Q(s,\theta) + \alpha \left[r + \max_{\theta'} Q\left(s', \theta'\right)\right]
\label{eq:QL}
\end{equation}

To that end, it learns the optimal state-action value function $Q^*(s,\theta) = r(s,\theta)+\mathbb{E}_{s'}\left(V^*(s')\right)$.
$Q$-learning is a \emph{stochastic approximation} process: it repeats the operation of Equation \ref{eq:QL} in all states and actions until convergence to $Q^*$ (which boils down to solving Equation \ref{eq:VI}). The optimal policy is then the greedy policy $\pi^*(s) = \arg\max_\theta Q^*(s,\theta)$. 

\section{Discussion}
The approach presented above generalizes straightforwardly to other problems and algorithms. Our goal in this contribution was to illustrate how a RL perspective on (optimal) Parameter Control can help bring new contributions to the Optimization field.
Extending this contribution to a larger class of problems opens new challenges:
\begin{itemize}
	\item Continuous actions (parameters) are a common limitation in RL, generally overcome using Policy Gradient methods.
	\item The state of an optimization process is problem and algorithm specific and might not always define a Markov process, thus leading to partial observability and/or approximations.
	\item The curse or dimensionality is a crucial issue in RL and introducing expert knowledge in the learning process can greatly help the convergence.
	\item Convergence to an optimal parameter control policy can take advantage of  sampling the optimization process at will.
	\item Minimizing the expected termination time is not the only relevant criterion. For instance, a natural alternative would be to maximize the time-discounted value function improvements (an approach close to the idea of \emph{regret minimization}).
\end{itemize}

\bibliographystyle{apalike}
\bibliography{paco_first}

\end{document}